\pdfoutput=1
\documentclass{article}
\usepackage{spconf,amsmath,graphicx,hyperref, amsfonts, bbm, tikz, amsthm}
\usetikzlibrary{calc}
\usepackage{tikz}
\usetikzlibrary{arrows.meta}
\usepackage{xcolor}
\definecolor{u1}{RGB}{0,114,178}
\definecolor{u2}{RGB}{213,94,0}
\definecolor{u3}{RGB}{0,158,115}
\definecolor{hd}{HTML}{6B7280}
\ninept
\usepackage{mathtools}
\mathtoolsset{showonlyrefs}
\usepackage{enumitem}
\usepackage{cleveref}
\hypersetup{hidelinks}
\usepackage{booktabs}

\def\x{{\mathbf x}} 
 
\def\h{{\mathbf h}} 
\def\X{{\mathbf X}} 
\def\y{{\mathbf y}} 
\def\w{{\mathbf w}} 
\def\W{{\mathbf W}} 
\def\b{{\mathbf b}} 
\def\z{{\mathbf z}}

\def\R{{\mathbb R}} 
\def\I{{\mathbf I}}

\def\D{{\mathbf D}}

\def\btheta{{\boldsymbol \theta}} 
\def\bTheta{{\boldsymbol \Theta}}

\newcommand{\dH}{d_H}

\newtheorem{theorem}{Theorem}

\newtheorem{proposition}[theorem]{Proposition}

\theoremstyle{remark}

\title{Bias-Only Training Dynamics in ReLU Networks}
\name{Shreyas Chaudhari, Jos\'e M. F. Moura}
\address{Department of Electrical and Computer Engineering, Carnegie Mellon University}
\begin{document}
\maketitle
\begin{abstract}
Neural networks with randomly initialized, fixed weights and trainable biases are universal approximators, yet the dynamics of bias-only training remain largely unexplored. We study these dynamics in one-hidden-layer ReLU networks through activation codes, the binary patterns of active neurons that partition the input space into convex polyhedral regions. We show that the gradients with respect to the biases depend on each input only through its activation code, and hence the training dynamics can be characterized with these codes. In particular, as long as the activation codes remain fixed, training reduces to a linear least-squares problem with the codes as features. Training can then be understood as a sequence of least-squares problems, with transitions occurring when neurons change their activation states. We show that the fraction of switching neurons vanishes as the network widens, and the training dynamics approach those of linear least squares on the initial activation codes in the infinite-width limit. Our experiments support these theoretical findings, showing that while activation switches can noticeably affect training in narrower networks, their relative frequency and the associated changes in the dynamics diminish as the width increases.
\end{abstract}

\section{Introduction}
\label{sec:intro}
Characterizing the training dynamics of neural networks is a central step
toward explaining how they transform data into useful representations. In
wide networks, these dynamics are described by the neural tangent kernel
(NTK)~\cite{jacot2018neural}, under which learning reduces to kernel
regression~\cite{lee2019wide, arora2019exact, chaudhari2025peer}. These
results concern training all weights and biases, yet recent work shows that
the biases alone carry substantial expressive power. Neural networks with random,
frozen weights and learned biases are universal
approximators~\cite{williams2025expressivity}, and bias-only fine-tuning is
competitive for adapting large models~\cite{zaken2022bitfit}. The dynamics of bias-only training, however, remain largely unexplored.

We study bias-only training of ReLU networks and show that its dynamics are
governed by the inherently combinatorial structure of these networks. A ReLU network is
a continuous piecewise affine function that partitions its input space into
convex polyhedra, each indexed by an \emph{activation code}, the binary vector
of neurons active at an input. The number of codes is related to
expressivity~\cite{montufar2014number, hanin2019deep}, and their distribution
under random inputs determines the distribution of the affine functions a
trained network realizes~\cite{chaudhari2025relu}. We show that these codes
also govern training: the gradient of the network output with respect to the
biases is a signed, scaled copy of the activation code, so each input affects training
only through its code. Consequently, the NTK is exactly a kernel on activation
codes, and two inputs are treated similarly to the extent that their codes
agree.

\textbf{Contributions.} We show that for ReLU networks trained only through
their biases, the NTK is \emph{exactly} a linear kernel on activation codes at
any width and throughout training, and that it changes only when neurons
flip (\Cref{prop:code-kernel}). Without flips, training is exactly least squares
on the codes, and it can fit arbitrary training targets if and only if the code
matrix has full row rank (\Cref{prop:frozen}). The squared distance that the NTK
induces between inputs is proportional to the Hamming distance between their
codes, which bounds how
differently two inputs evolve during training (\Cref{prop:alike}). At
initialization, the codes are random-hyperplane hashes, and the kernel converges
to a scaled collision probability, namely the arc-cosine kernel~\cite{cho2009kernel}
(\Cref{prop:random-codes}). Finally, we show that
at most $O(\sqrt{m})$ neurons flip per input, yet their fraction and the
resulting kernel drift both shrink as $O(m^{-1/2})$ (\Cref{prop:travel}), as
confirmed by simulations.

\textbf{Related work.} Bounding the number of ReLU units that change state during training is central to over-parameterized convergence proofs~\cite{du2019gradient, arora2019fine, allenzhu2019convergence, oymak2020toward}, while finite-width NTK corrections~\cite{hanin2020finite}, activation flips under large bias initialization~\cite{yang2024sparse}, and hidden-feature drift under lazy training~\cite{starnaud2025revisiting} have been studied. The expressivity and practical benefits of bias-only and other partial training have been studied~\cite{williams2025expressivity, zaken2022bitfit, frankle2021training}, but the training dynamics of such settings have received comparatively little attention. Training only the output layer of a random network yields a random-feature model~\cite{rahimi2007random}; bias-only training instead moves the hidden features themselves. The geometry of ReLU networks has been studied through their number of regions, polyhedral structure, and distribution~\cite{montufar2014number, balestriero2018spline, gaines2026characterizing, chaudhari2025relu}, as well as through kernels defined by gating patterns and neural paths~\cite{fiat2019decoupling, lakshminarayanan2020neural}. Hamming distances between random-hyperplane codes approximate angles~\cite{goemans1995improved, charikar2002similarity}, and random (biased) tessellations preserve and separate data~\cite{plan2014dimension, dirksen2022separation}. In this paper, we connect the geometry of activation codes in ReLU networks directly to the bias-only training regime.

\section{Preliminaries}
\label{sec:prelim}

\subsection{Lazy Training}
\label{sec:lazy_training}
Consider training a neural network $f(\x;\btheta)$ with
parameters $\btheta$ on a dataset $\{(\x_i,y_i)\}_{i=1}^n\subset\R^d\times\R$, with
inputs $\X=[\x_1,\!\ldots\!,\x_n]^\top\in\R^{n\times d}$ and targets
$\y=[y_1,\ldots,y_n]^\top\in\R^n$. The network is trained by gradient flow, $\dot\btheta_t=-\nabla_{\btheta}\mathcal L(\btheta_t)$, on the squared loss
\begin{equation}
\mathcal L(\btheta)=\tfrac{1}{2}\left\|f(\X;\btheta)-\y\right\|_2^2
\label{eq:loss}
\end{equation}
where $f(\X;\btheta)\in\R^n$ collects the outputs. Denoting $f_t(\x)=f(\x;\btheta_t)$, the chain rule yields the dynamics
\begin{align}
\dot \btheta_t &= - \left(\frac{\partial f(\X; \btheta_t)}{\partial \btheta_t}\right)^\top (f(\X; \btheta_t) - \y) \label{eq:ntk-param-flow} \\
\dot f_t(\x) &= -\,\Theta_t(\x,\X)\,\left(f_t(\X)-\y\right)\label{eq:ntk-flow}
\end{align}
where
$\Theta_t(\x,\x')=\nabla_{\btheta}f(\x;\btheta_t)^\top\nabla_{\btheta}f(\x';\btheta_t)$
is the \emph{neural tangent kernel} (NTK) \cite{jacot2018neural},
$\Theta_t(\x,\X)\in\R^{1\times n}$ collects its values on the training
inputs, and $\bTheta_t\in\R^{n\times n}$, with
$(\bTheta_t)_{ij}=\Theta_t(\x_i,\x_j)$, is its Gram matrix. In general, $\Theta_t$ evolves with $\btheta_t$, so \eqref{eq:ntk-flow} is
challenging to solve analytically. As shown in~\cite{jacot2018neural,chizat2019lazy,lee2019wide,liu2020linearity}, however, the parameters of a sufficiently wide network stay close to $\btheta_0$ throughout training, and the network can be well approximated by its linearization around its initial parameters $\btheta_0$:
\begin{align}
f(\x;\btheta_t)\approx f(\x;\btheta_0)
+\nabla_{\btheta}f(\x;\btheta_0)^\top(\btheta_t-\btheta_0).
\label{eq:linearization}
\end{align}
In this setting, the kernel is approximately constant,
$\Theta_t\approx\Theta_0$, and the dynamics in~\eqref{eq:ntk-param-flow} and~\eqref{eq:ntk-flow} can be solved~\cite{lee2019wide,chaudhari2025peer}:
\begin{align}
\btheta_t&=\btheta_0+\left(\frac{\partial f(\X;\btheta_0)}{\partial\btheta_0}\right)^{\top}
\bTheta_0^{-1}\left(\I-e^{-t\bTheta_0}\right)\left(\y-f_0(\X)\right)\\
f_t(\x)&=f_0(\x)+\Theta_0(\x,\X)\,\bTheta_0^{-1}\big(\I-e^{-t\bTheta_0}\big)\left(\y-f_0(\X)\right)
\label{eq:kgf}
\end{align}

\subsection{ReLU Networks and Activation Codes}
Let $f:\R^d\to\R$ be a one-hidden-layer network with~$m$ units, weight
scale $s_w>0$, and bias scale $s_b>0$,
\begin{align}
\h(\x) &= \frac{s_w}{\sqrt{d}}\,\W_1\x + s_b\,\b,\;
\qquad f(\x) = \frac{s_w}{\sqrt{m}}\,\w_{2}^\top\sigma\left(\h(\x)\right)
\end{align}
where $\sigma(a)=\max(a,0)$ is the ReLU activation applied elementwise. The entries of $\W_1$, whose $j$th row we denote $\w_{1,j}^\top$, and of $\b$
are initialized i.i.d.\ $\mathcal N(0,1)$, and those of $\w_{2}$ are i.i.d.\ uniform on $\{\pm1\}$. We train only the bias $\b$, keeping the weights fixed at their random initial values. This parameterization is consistent with the standard infinite-width scaling under which the NTK admits a well-defined limit as $m\to\infty$ \cite{jacot2018neural, lee2019wide}.
\begin{figure}[htpb]
\centering
\colorlet{hd0}{hd!4}
\colorlet{hd1}{hd!25}
\colorlet{hd2}{hd!48}
\colorlet{hd3}{hd!72}
\begin{tikzpicture}[x=0.72cm,y=0.72cm,
  hyp/.style={line width=1.1pt},
  act/.style={-{Stealth[length=4.5pt,width=4pt]},line width=1pt},
  code/.style={font=\scriptsize\ttfamily,text=black,fill=white,fill opacity=0.8,
               text opacity=1,inner sep=1pt,rounded corners=1pt},
  lab/.style={font=\scriptsize},
  cross/.style={circle,draw=white,line width=0.6pt,inner sep=1.5pt},
  pt/.style={circle,fill=black,draw=white,line width=0.6pt,inner sep=1.6pt}]
  \fill[hd1] (0.000,0.000) -- (1.500,0.000) -- (1.912,1.646) -- (0.000,1.200) -- cycle;
  \fill[hd3] (2.250,3.000) -- (2.550,4.200) -- (1.350,4.200) -- cycle;
  \fill[hd2] (1.912,1.646) -- (2.250,3.000) -- (1.350,4.200) -- (0.000,4.200) -- (0.000,1.200) -- cycle;
  \fill[hd1] (4.500,0.000) -- (6.000,0.000) -- (6.000,2.600) -- (3.064,1.915) -- cycle;
  \fill[hd0] (1.500,0.000) -- (4.500,0.000) -- (3.064,1.915) -- (1.912,1.646) -- cycle;
  \fill[hd2] (6.000,2.600) -- (6.000,4.200) -- (2.550,4.200) -- (2.250,3.000) -- (3.064,1.915) -- cycle;
  \fill[hd1] (2.250,3.000) -- (1.912,1.646) -- (3.064,1.915) -- cycle;
  \draw[hyp,u1] (1.5,0) -- (2.55,4.2);
  \draw[hyp,u2] (0,1.2) -- (6,2.6);
  \draw[hyp,u3] (4.5,0) -- (1.35,4.2);
  \node[lab,u1,below] at (1.5,0) {$h_1(\x)=0$};
  \node[lab,u3,below] at (4.5,0) {$h_3(\x)=0$};
  \node[lab,u2,anchor=east] at (-0.05,1.2) {$h_2(\x)=0$};
  \draw[act,u1] (1.725,0.9) -- ++(0.36,-0.09);
  \draw[act,u2] (1.3,1.503) -- ++(-0.085,0.36);
  \draw[act,u3] (4.2,0.4) -- ++(-0.29,-0.22);
  \foreach \p/\c in {(0.6,0.55)/001,(2.3,0.4)/101,(5.25,0.65)/100,(4.6,3.65)/110,
                     (0.6,3.0)/011,(1.95,3.9)/010,(2.41,2.19)/111}
    \node[code] at \p {\c};
  \draw[black!35] (0,0) rectangle (6,4.2);
  \draw[black!75,densely dashed,line width=0.9pt] (3.3,0.8) -- (5.3,3.0);
  \node[cross,fill=u3] at (3.629,1.162) {};
  \node[cross,fill=u2] at (4.65,2.285) {};
  \node[pt,label={[font=\small,inner sep=1pt]below left:$\x$}] at (3.3,0.8) {};
  \node[pt,label={[font=\small,inner sep=1pt]above left:$\x'$}] at (5.3,3.0) {};
  \node[anchor=north west,font=\small] at (6.25,4.15) {$\begin{aligned}
    \z(\x)  &= \textcolor{u1}{1}\,\textcolor{u2}{\mathbf 0}\,\textcolor{u3}{\mathbf 1}\\
    \z(\x') &= \textcolor{u1}{1}\,\textcolor{u2}{\mathbf 1}\,\textcolor{u3}{\mathbf 0}\\[2pt]
    \dH(\x,\x') &= 2
  \end{aligned}$};
  \node[font=\scriptsize,anchor=south west,inner sep=0pt] at (6.45,0.75)
    {$\dH(\,\cdot\,,\x)$};
  \foreach \d in {0,1,2,3} {
    \fill[hd\d] (6.45+0.6*\d,0.2) rectangle ++(0.5,0.4);
    \draw[black!35] (6.45+0.6*\d,0.2) rectangle ++(0.5,0.4);
    \node[font=\scriptsize] at (6.7+0.6*\d,-0.1) {\d};
  }
\end{tikzpicture}
\caption{Activation regions of a one-hidden-layer network with $m=3$ neurons on
a 2-D input. Neuron $j$ is active on the side of its boundary
$h_j(\x)=0$ marked by the arrow. Regions are labeled by their codes
$z_1z_2z_3$ and shaded by their Hamming distance to $\z(\x)$. The path from
$\x$ to $\x'$ crosses 2 hyperplanes, so $\dH(\x,\x')=2$.}
\label{fig:regions}
\end{figure}
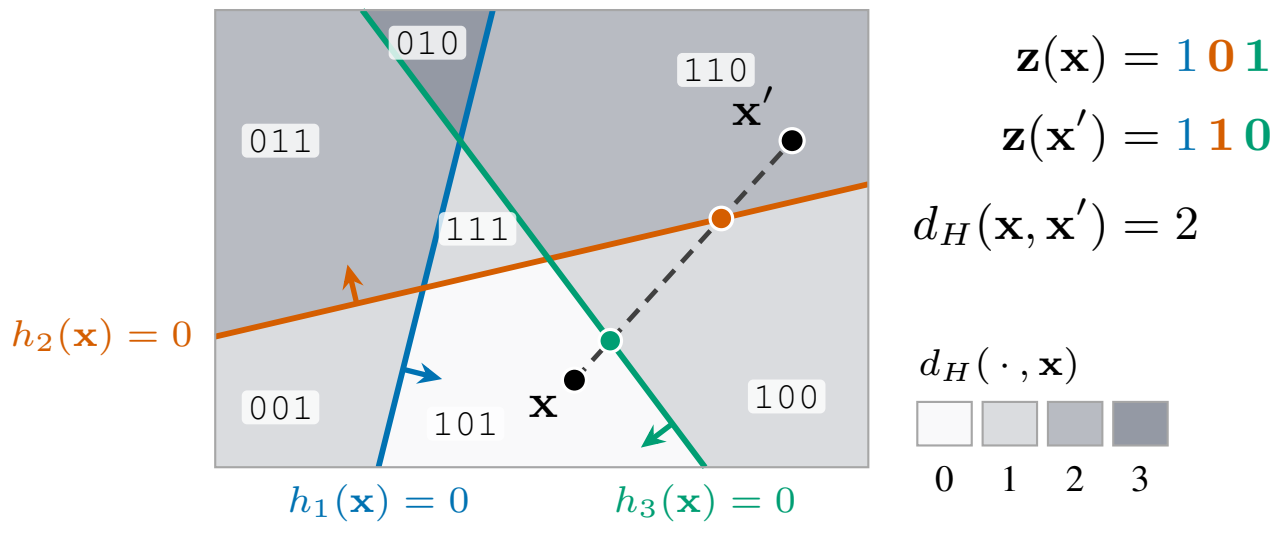

The \emph{activation code} is defined as
\begin{align}
    \z(\x) = \mathbbm{1}\left[\h(\x)>0\right] \in \{0, 1\}^m
\end{align}
and records which ReLU units are active in response to an input $\x$. These activations are related to the inherent geometry of ReLU networks. ReLU networks are continuous piecewise affine functions over convex polyhedral partitions of the input space. Two inputs with the same full activation code belong to the same linear region. For two inputs $\x,\x'$, the Hamming distance (number of bit differences) between their activation codes is
\begin{align}
\dH(\x,\x') = \#\{j : z_j(\x) \neq z_j(\x')\} = \|\z(\x) - \z(\x')\|_2^2. \label{eq:hamming_dist}
\end{align}
Geometrically, the Hamming distance counts the neuron hyperplanes $h_j(\x)=0$ that separate $\x$ and $\x'$, i.e., those crossed by the segment between them, as illustrated in~\Cref{fig:regions}. 
\section{Bias-Only Training Dynamics}
\label{sec:hamming_geometry}

We study the training dynamics when only the bias $\b$ is
learned by gradient flow. Throughout, we use the notation $\mathbf r_t=f_t(\X)-\y$ for the training residual at time $t$, $c=s_b^2s_w^2$, and $\D(\x)=\mathrm{diag}(\z(\x))$.
The gradient of the network output with respect to the bias is
\begin{align}
\nabla_\b f(\x)
=\sqrt{\tfrac{c}{m}}\,\w_2\odot\z(\x).
\label{eq:bias-grad}
\end{align}
The gradient therefore depends on the input only through its
activation code and is piecewise constant with respect to the inputs.

\subsection{Finite Width: A Linear Kernel on Codes}
\label{subsec:finite}

Let $\mathbf Z=[\z(\x_1),\ldots,\z(\x_n)]^\top$ be the matrix
of training codes. Along a training trajectory, we write
$\z_t$ and $\Theta_t$ for the codes and kernel at time $t$, and
let $\tau_t(\x)=\|\z_t(\x)-\z_0(\x)\|_2^2$ be the number of bits
of the code of $\x$ that have flipped since initialization, so
that $\tau_t(\x)/m$ is the fraction of flipped bits.

\begin{proposition}[Linear code kernel]
\label{prop:code-kernel}
For all inputs $\x,\x'$:
\begin{enumerate}[label=(\roman*),leftmargin=*]
\setlength{\itemsep}{0pt}
\item
at any bias,
\[
\Theta(\x,\x')=\frac{c}{m}\,\z(\x)^\top\z(\x'),
\qquad
\bTheta=\frac{c}{m}\,\mathbf Z\mathbf Z^\top;
\]
\item
along any training trajectory, for all $t$,
\[
|\Theta_t(\x,\x')-\Theta_0(\x,\x')|
\le\frac{c}{m}\big(\tau_t(\x)+\tau_t(\x')\big).
\]
\end{enumerate}
\end{proposition}

\begin{proof}
(i) From \eqref{eq:bias-grad} it follows that
\[\Theta(\x,\x')=\tfrac{c}{m}\w_2^\top\D(\x)\D(\x')\w_2,\]
which equals $\tfrac{c}{m}\z(\x)^\top\z(\x')$ since
$w_{2,j}^2=1$.
(ii) Let $\z,\z'$ and $\z_0,\z_0'$ be the codes of $\x,\x'$
at times $t$ and $0$. Adding and subtracting $\z_0^\top\z'$
gives \[\tfrac{m}{c}(\Theta_t-\Theta_0)=(\z-\z_0)^\top\z'
+\z_0^\top(\z'-\z_0').\]
Since $\z-\z_0\in\{-1,0,1\}^m$ and $\z'\in\{0,1\}^m$, the
first term is at most $\|\z-\z_0\|_2^2=\tau_t(\x)$ in
magnitude, and likewise the second is at most $\tau_t(\x')$.
\end{proof}

\Cref{prop:code-kernel} shows that the bias-only NTK is
exactly a linear kernel on activation codes at any width. Specifically,
$\tfrac{m}{c}\Theta(\x,\x')$ counts the neurons active at both inputs,
and each code flip changes a kernel entry by at most $c/m$.

\begin{proposition}[Fixed codes]
\label{prop:frozen}
If the training codes remain fixed on $[0,T]$, bias-only
training is exactly least squares on the initial
codes, with predictions
\[
f_t(\X)=\y+e^{-t\bTheta_0}\big(f_0(\X)-\y\big),
\qquad t\in[0,T].
\]
If the codes remain fixed for all $t\ge0$, then
$f_t(\X)\to\y+\mathbf P \mathbf r_0$, where $\mathbf P$ is the
orthogonal projector onto the null space of $\mathbf Z^\top$. The residual $\mathbf P\mathbf r_0$ vanishes for every target if and only if $\mathbf Z$ has full row rank.
\end{proposition}

\begin{proof}
Since $\sigma(\h(\x))=\D(\x)\h(\x)$, the network output
splits as
\[
f(\x;\b)=\eta(\x)+\z(\x)^\top\boldsymbol\beta,
\qquad
\boldsymbol\beta=\sqrt{\tfrac{c}{m}}\,(\w_2\odot\b),
\]
where $\eta(\x)=\frac{s_w^2}{\sqrt{md}}\w_2^\top\D(\x)\W_1\x$
is the contribution of the fixed weights.
When the training codes remain fixed, the vector
$\boldsymbol\eta$ of values $\eta(\x_i)$ is constant, so
$f(\X;\b)=\boldsymbol\eta+\mathbf Z\boldsymbol\beta$ and the loss is
the least-squares objective
$\tfrac12\|\mathbf Z\boldsymbol\beta-(\y-\boldsymbol\eta)\|_2^2$,
with the binary codes as features.
Since $w_{2,j}^2=1$, gradient flow on $\b$ gives
$\dot{\boldsymbol\beta}_t=-\tfrac{c}{m}\mathbf Z^\top\mathbf r_t$,
which is gradient flow on this objective with time rescaled by $c/m$.
Hence $\dot{\mathbf r}_t=\mathbf Z\dot{\boldsymbol\beta}_t
=-\bTheta_0\mathbf r_t$ by \Cref{prop:code-kernel}(i).
Solving this linear differential equation yields
$\mathbf r_t=e^{-t\bTheta_0}\mathbf r_0$.
Since $\bTheta_0$ is symmetric positive semidefinite with
$\ker(\bTheta_0)=\ker(\mathbf Z^\top)$, $e^{-t\bTheta_0}\to\mathbf P$
as $t\to\infty$, and $\mathbf P=\mathbf 0$ if and only if
$\operatorname{rank}(\mathbf Z)=n$.
\end{proof}

With fixed codes, the network behaves as a gated linear model
\cite{fiat2019decoupling,lakshminarayanan2020neural} whose
training predictions lie in an affine subspace of dimension
$\operatorname{rank}(\mathbf Z)$. Fitting targets outside this space requires changes in activation codes, complementing the expressivity results of~\cite{williams2025expressivity}.

In general, activation codes can change during training.
Neuron $j$ is active at $\x_i$ when
$b_j>\gamma_{ij}=-\tfrac{s_w}{s_b\sqrt d}\,\w_{1,j}^\top\x_i$.
These thresholds partition each coordinate of the bias vector
into intervals with fixed activation states, and together they
form an axis-aligned grid in bias space.
Within each cell $k$, the training codes $\mathbf Z_k$ remain
fixed, and so does $\boldsymbol\eta_k$, the vector of values $\eta(\x_i)$.
The loss is then the least-squares objective
$\tfrac12\|\mathbf Z_k\boldsymbol\beta-(\y-\boldsymbol\eta_k)\|_2^2$
with features $\mathbf Z_k$. Consequently, gradient flow follows
\emph{piecewise-linear} dynamics. Within cell $k$, the residual
satisfies $\dot{\mathbf r}_t=-\bTheta_k\mathbf r_t$, where
$\bTheta_k=\tfrac{c}{m}\mathbf Z_k\mathbf Z_k^\top$.
When an entry of the bias crosses a threshold, the corresponding activation bit flips, so the
dynamics switch to those of the neighboring cell. Bias-only
training thus proceeds through a sequence of least-squares
problems, and \Cref{prop:frozen} describes the special case in
which no threshold is crossed.

\subsection{Linear Code Kernel Induces Hamming Geometry}
\label{subsec:hamming}

Recall that the Hamming distance \eqref{eq:hamming_dist}
counts the neuron hyperplanes separating two inputs
(\Cref{fig:regions}).
By \eqref{eq:bias-grad}, and since $w_{2,j}^2=1$,
\begin{align}
\|\nabla_\b f(\x)-\nabla_\b f(\x')\|_2^2
=\tfrac{c}{m}\dH(\x,\x').
\label{eq:hamming}
\end{align}
The Hamming distance hence measures how differently two inputs
respond to bias updates. Expanding \eqref{eq:hamming_dist} with
\Cref{prop:code-kernel}(i) also gives
\begin{align}
\Theta(\x,\x')=\tfrac{c}{2m}\big(\|\z(\x)\|_2^2+\|\z(\x')\|_2^2
-\dH(\x,\x')\big),
\label{eq:offset}
\end{align}
so the kernel decreases with Hamming distance, offset by the
number of neurons active at each input. The following
proposition relates this geometry to the training dynamics.

\begin{proposition}[Hamming-close inputs train alike]
\label{prop:alike}
Let the Hamming distance at time $t$ be
$d_{H,t}(\x,\x')=\|\z_t(\x)-\z_t(\x')\|_2^2$. Under gradient flow, for
almost every $t$,
\[
\left|\tfrac{d}{dt}[f_t(\x)-f_t(\x')]\right|
\le \tfrac{c}{m}d_{H,t}(\x,\x')\,\|\mathbf r_t\|_1.
\]
\end{proposition}

\begin{proof}
By the NTK dynamics,
\[
\tfrac{d}{dt}[f_t(\x)-f_t(\x')]
=-\sum_{i=1}^n
[\Theta_t(\x,\x_i)-\Theta_t(\x',\x_i)]\,r_{t,i}.
\]
By \Cref{prop:code-kernel}(i), each kernel difference is
$\tfrac{c}{m}\sum_{j}\big(z_{t,j}(\x)-z_{t,j}(\x')\big)z_{t,j}(\x_i)$.
Since $z_{t,j}(\x_i)\in\{0,1\}$, each term is at most
$|z_{t,j}(\x)-z_{t,j}(\x')|$ in magnitude, which equals $1$ if
the two inputs differ in bit $j$ and $0$ otherwise. Hence
$|\Theta_t(\x,\x_i)-\Theta_t(\x',\x_i)|\le\tfrac{c}{m}d_{H,t}(\x,\x')$.
Applying the triangle inequality to the sum and this bound to
each term gives the result, since $\sum_i|r_{t,i}|=\|\mathbf r_t\|_1$.
\end{proof}

Thus inputs with similar codes receive similar updates, and
inputs with identical codes receive the same updates, so their
prediction difference stays constant while their codes
coincide, however far apart they are in input space.

\subsection{Infinite Width: Random Hyperplane Codes}
\label{subsec:infinite}

The preceding results hold for arbitrary biases and arbitrary network widths.
We now consider the infinite-width limit $(m \to \infty)$ at initialization.
Let $\mathbf g_j=(\w_{1,j},b_j)\sim
\mathcal N(\mathbf 0,\mathbf I_{d+1})$ and
$\bar\x=(s_w\x/\sqrt d,s_b)$, so that
$h_j(\x)=\mathbf g_j^\top\bar\x$.
Additionally, we denote by $\phi$ the angle between $\bar\x$ and $\bar\x'$.

\begin{proposition}[Random hyperplane codes]
\label{prop:random-codes}
Let $\Theta^\star(\x,\x')=\tfrac{c}{2}(1-\phi/\pi)$.
At initialization, for any fixed $\x,\x'$:
\begin{enumerate}[label=(\roman*),leftmargin=*]
\setlength{\itemsep}{0pt}
\item $\mathbb E[\dH(\x,\x')]/m=\phi/\pi$;
\item $\mathbb E[\Theta_0(\x,\x')]=\Theta^\star(\x,\x')$;
\item with probability at least $1-\delta$,
$|\Theta_0(\x,\x')-\Theta^\star(\x,\x')|
\le c\sqrt{\log(2/\delta)/(2m)}$.
\end{enumerate}
\end{proposition}

\begin{proof}
Each neuron defines a random hyperplane separating
$\bar\x$ and $\bar\x'$ with probability $\phi/\pi$
\cite{goemans1995improved,charikar2002similarity},
giving (i). By symmetry, both inputs are active with
probability $(1-\phi/\pi)/2$, a Gaussian orthant
probability~\cite{chaudhari2025relu}. Taking expectations in
$\Theta_0(\x,\x')=\tfrac{c}{m}\sum_j z_j(\x)z_j(\x')$
gives (ii). Since the summands are independent and bounded,
Hoeffding's inequality gives
$\Pr(|\Theta_0-\Theta^\star|\ge\varepsilon)\le 2e^{-2m\varepsilon^2/c^2}$,
which implies (iii).
\end{proof}

At initialization, the angle between lifted inputs thus sets
the expected Hamming distance between their codes. Inputs at a
small angle share most bits and train alike, while inputs at a
large angle are weakly coupled. As the width grows, the kernel
concentrates around $\Theta^\star$, a scaled degree-zero
arc-cosine kernel on the lifted inputs~\cite{cho2009kernel}.

\subsection{How Wide Is Wide Enough?}
\label{sec:travel}

By \Cref{prop:code-kernel}(ii), the kernel changes only
when activation codes change. Increasing width does not
eliminate code flips but reduces their fraction.

\begin{proposition}[Code flips diminish with width]
\label{prop:travel}
Suppose $\|\x_i\|_2=1$ for all training inputs, and fix
a training horizon $T$ and $\delta\in(0,1)$.
There exists a constant $K$, independent of $m$, such
that with probability at least $1-\delta$, for all
$t\in[0,T]$ and all $i,j$,
\[
\tau_t(\x_i)\le K\sqrt m,
\qquad
|\Theta_t(\x_i,\x_j)-\Theta_0(\x_i,\x_j)|
\le\frac{2cK}{\sqrt m}.
\]
\end{proposition}

\begin{proof}
Following~\cite{du2019gradient,yang2024sparse},
we bound the bias movement, code flips, and kernel drift.

\emph{Bias movement.}
Gradient flow does not increase the loss, so
$\|\mathbf r_t\|_2\le\|\mathbf r_0\|_2$.
At initialization,
$\mathbb E[f_0(\x_i)^2]=s_w^2q/2$, where
$q=\|\bar\x_i\|_2^2=s_w^2/d+s_b^2$, so
$\mathbb E\|\mathbf r_0\|_2^2$ is independent of $m$.
By Markov's inequality, there exists $B$, independent
of $m$, such that $\|\mathbf r_0\|_2\le B$ with
probability at least $1-\delta/2$.
On this event, \eqref{eq:bias-grad} gives
\[|\dot b_{t,j}|\le\sqrt{c/m}\,\|\mathbf r_t\|_1\le B\sqrt{cn/m}.\]
Each bias therefore moves by at most
$R=TB\sqrt{cn/m}$ over $[0,T]$.

\emph{Code flips.}
A neuron can change state at $\x_i$ only if
$|h_{0,j}(\x_i)|\le s_bR$.
Since $h_{0,j}(\x_i)\sim\mathcal N(0,q)$ with
$q\ge s_b^2$, this event has probability at most
$\sqrt{2/\pi}R=O(m^{-1/2})$.
Hence the expected number of neurons close enough
to flip at each input is $O(\sqrt m)$.
Markov's inequality and a union bound over the
$n$ training inputs imply that, with probability
at least $1-\delta/2$, this number is at most
$K\sqrt m$ for every input, for a suitable
constant $K$ independent of $m$.
Consequently, $\tau_t(\x_i)\le K\sqrt m$
for all $t\le T$.

\emph{Kernel drift.}
By \Cref{prop:code-kernel}(ii),
\[|\Theta_t(\x_i,\x_j)-\Theta_0(\x_i,\x_j)|
\le\tfrac{c}{m}(\tau_t(\x_i)+\tau_t(\x_j))\le 2cK/\sqrt m.\]
A union bound completes the proof.
\end{proof}

Although the number of code flips may grow
with width, their fraction and the kernel drift
vanish as $O(m^{-1/2})$ with high probability. With
\Cref{prop:random-codes}, the kernel thus approaches
$\Theta^\star$ over any fixed horizon, recovering the
lazy-training regime.
\section{Experiments}
\label{sec:experiments}

We train the biases of a one-hidden-layer network with
$s_w^2=2$ and $s_b=1$ (so $c=2$), initialized as in
\Cref{sec:prelim}, using full-batch gradient descent with
step size $1/\lambda_{\max}(\bTheta_0)$.
The $n=30$ training inputs and $500$ test inputs are drawn
uniformly from the unit sphere in $\R^5$, with targets
$y=\sin(2x_1)+\frac12x_2^2$. To study nearby inputs, we
create a perturbed version of each of $100$ test inputs at a
distance between $10^{-3}$ and $10^{-1}$. We train for $3000$
steps with network widths $m\in\{256,1024,4096,16384\}$.

\begin{figure}[t]
\centering
\begin{minipage}[b]{0.49\linewidth}
  \centering
  \includegraphics[width=\linewidth]{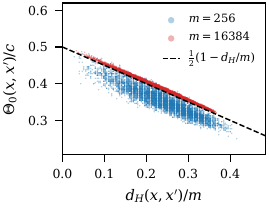}
  \\[-1pt]{\small (a)}
\end{minipage}\hfill
\begin{minipage}[b]{0.49\linewidth}
  \centering
  \includegraphics[width=\linewidth]{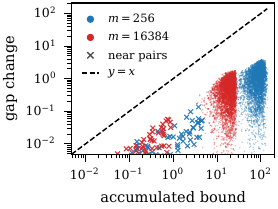}
  \\[-1pt]{\small (b)}
\end{minipage}
\\[1pt]
\begin{minipage}[b]{0.49\linewidth}
  \centering
  \includegraphics[width=\linewidth]{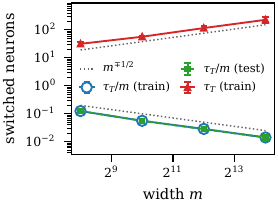}
  \\[-1pt]{\small (c)}
\end{minipage}\hfill
\begin{minipage}[b]{0.49\linewidth}
  \centering
  \includegraphics[width=\linewidth]{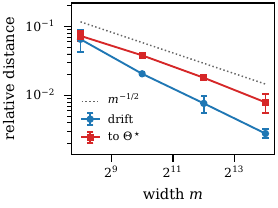}
  \\[-1pt]{\small (d)}
\end{minipage}
\vspace{-1pt}
\caption{
(a) Initial kernel $\Theta_0/c$ versus normalized Hamming
distance $\dH/m$, approaching $\frac12(1-\dH/m)$ with width.
(b) Change in prediction differences versus the bound of
\Cref{prop:alike}; crosses mark nearby pairs, and the dashed line
marks equality.
(c) Number $\tau_T$ and fraction $\tau_T/m$ of flipped bits
versus width. Training and test fractions nearly coincide;
dotted lines show $m^{1/2}$ and $m^{-1/2}$ scaling.
(d) Kernel drift and distance to $\Theta^\star$ versus
width, with $m^{-1/2}$ scaling dotted.
}
\label{fig:experiments}
\end{figure}

\Cref{fig:experiments}(a) plots the initial kernel against
the Hamming distance between test inputs. By \eqref{eq:offset},
\[
\frac{\Theta_0}{c}=\frac12\Big(1-\frac{\dH}{m}\Big)
+\frac{\|\z_0(\x)\|_2^2+\|\z_0(\x')\|_2^2-m}{2m},
\]
where $\|\z_0(\x)\|_2^2$ counts the neurons active at $\x$.
The second term has mean zero and is of order $m^{-1/2}$.
At $m=256$, only $43\%$ of the biases are positive, so most
points lie slightly below $\frac12(1-\dH/m)$. As width
increases, the points concentrate around this line, as
predicted by \Cref{prop:random-codes}.

\Cref{fig:experiments}(b) tests \Cref{prop:alike} using all
pairs among the $100$ selected test inputs and their nearby
copies. For each pair, the vertical axis shows the change in
prediction difference over training, namely
$|[f_T(\x)-f_T(\x')]-[f_0(\x)-f_0(\x')]|$, while the horizontal
axis shows the bound from \Cref{prop:alike}, summed over
training steps and multiplied by the step size. We observe that
nearby input pairs (crosses) have similar activation codes and
small changes in prediction difference, with all observed
changes below the bound (dashed diagonal).

\Cref{fig:experiments}(c) and the first three columns of
\Cref{tab:flips} test \Cref{prop:travel}. As the width increases
from $m=256$ to $m=16384$, the number $\tau_T$ of flipped bits
per training input rises from $32$ to $231$, while the fraction
$\tau_T/m$ decreases from $0.125$ to $0.014$ (blue circles).
The fitted exponents, $+0.48$ and $-0.52$, are consistent with
the $O(\sqrt m)$ and $O(m^{-1/2})$ bounds of \Cref{prop:travel}.
The fraction of flipped bits at test inputs (green squares)
closely matches that at training inputs. Each flip changes a
kernel entry by at most $c/m$, so the decreasing fraction of
flipped bits keeps the kernel drift small.

The kernel drift decreases from $0.065$ to $0.0028$ as the width
increases (\Cref{fig:experiments}(d)), with a fitted slope of
$-0.75$, faster than the $m^{-1/2}$ bound of \Cref{prop:travel}.
The bound accounts for every flip, although a flip at $\x$
affects $\Theta(\x,\x')$ only when the neuron is also active at
$\x'$, and flips in opposite directions can cancel. As the width
increases, the trained network also approaches the fixed-code
predictor of \Cref{prop:frozen}, which keeps all activation codes
at their initial values. The difference between their test
predictions, relative to the change in the fixed-code predictor
from initialization, decreases from $17\%$ to $1.5\%$, while the
training loss falls from $10^{-2.2}$ to $10^{-7.5}$
(\Cref{tab:flips}). Thus, even though activation codes continue
to change, the training dynamics of wide networks are well
described by least squares on their initial codes.

\begin{table}[!t]
\centering
\caption{Flipped code bits per input at the end of training $T$:
fraction $\tau_T/m$ and count $\tau_T$ at training inputs, and
fraction at test inputs. Kernel columns show relative drift
$\|\bTheta_T-\bTheta_0\|_F/\|\bTheta_0\|_F$ and distance
$\|\bTheta_0-\bTheta^\star\|_F/\|\bTheta^\star\|_F$. Fit columns
report final training loss and test difference from the
fixed-code least-squares (LS) predictor of \Cref{prop:frozen},
relative to its change from initialization. The final row gives
fitted log-log slopes in $m$.}
\label{tab:flips}
\footnotesize
\setlength{\tabcolsep}{2.6pt}
\begin{tabular}{@{}r ccc cc cc@{}}
\toprule
 & \multicolumn{3}{c}{Flipped bits} & \multicolumn{2}{c}{Kernel}
 & \multicolumn{2}{c}{Fit} \\
\cmidrule(lr){2-4}\cmidrule(lr){5-6}\cmidrule(l){7-8}
$m$ & $\tau_T/m$ & $\tau_T$ & test & drift & to $\Theta^\star$
 & $\log_{10}$ loss & vs.\ LS \\
\midrule
256   & 0.125 & 32  & 0.125 & 0.065  & 0.073  & $-2.2$ & 0.17  \\
1024  & 0.056 & 57  & 0.056 & 0.021  & 0.038  & $-5.3$ & 0.066 \\
4096  & 0.029 & 117 & 0.028 & 0.0077 & 0.018  & $-6.3$ & 0.021 \\
16384 & 0.014 & 231 & 0.014 & 0.0028 & 0.0079 & $-7.5$ & 0.015 \\
\midrule
slope & $-0.52$ & $+0.48$ & $-0.52$ & $-0.75$ & $-0.53$ & -- & $-0.60$ \\
\bottomrule
\end{tabular}
\end{table}

\section{Conclusion}
\label{sec:conclusion}
We have shown that when only the biases of a ReLU network are trained, the
training dynamics are governed by its binary activation codes. The NTK is a
linear kernel on these codes at any width, and inputs with similar codes
undergo similar changes in their predictions. Between changes in activation
states, training is linear least squares on the codes. As the network width
increases, the number of activation switches per input grows at most as
$O(\sqrt m)$, while their fraction vanishes. As a result, the kernel drift
decreases as $O(m^{-1/2})$, and the kernel approaches its infinite-width
limit. Together, these results connect the discrete structure of the regions
formed by ReLU activations with the continuous dynamics of training. Future
work may study extensions to deeper networks.

\newpage 
\bibliographystyle{IEEEbib}
\bibliography{refs}

\end{document}